# SHOOTOUT-89, A Comparative Evaluation of Knowledge-based Systems that Forecast Severe Weather[1]

William R. Moninger[2]

John A. Flueck[3]

Cynthia Lusk[4]

William F. Roberts[3]

## 0. ABSTRACT

During the summer of 1989, the Forecast Systems Laboratory of the National Oceanic and Atmospheric Administration sponsored an evaluation of AI systems that forecast severe convective storms. The evaluation experiment, called SHOOTOUT-89, took place in Boulder, CO, and focussed on storms over the northeastern Colorado plains.

Several systems participated in SHOOTOUT-89. These systems include traditional expert systems, an analogy-based system, and a system developed using methods from the cognitive science/judgment analysis tradition.

Each day of the exercise, the systems forecasted for each of four specified forecast regions in northeastern Colorado, the probabilities of occurrence of: nonsignificant weather, significant weather, and severe weather. A verification coordinator working at the Denver Weather Service Forecast Office gathered ground-truth data from a network of observers.

Systems were evaluated on forecast skill, and also on other metrics such as timeliness, ease of learning, ease of use, degree of portability to other locations. We report here initial results of the SHOOTOUT-89 experiment.

## 1. INTRODUCTION

During the summer of 1989, the Forecast Systems Laboratory (FSL) of the National Oceanic and Atmospheric Administration is sponsoring an evaluation of Artificial Intelligence (AI) systems that forecast convective storms. The evaluation exercise, called SHOOTOUT-89, is taking place in Boulder, CO, and focuses on storms over the northeastern Colorado plains.

A comparative study of AI systems was first proposed at the second workshop on Artificial Intelligence Research in Environmental Science (AIRIES) in 1987. At that time, it was noted that several AI systems had been devised that produce forecasts of severe convective storms, and that the time might be appropriate to bring the systems together for a quasi-operational intercomparison. The location chosen was the high plains and foothills of northeastern Colorado. This region is suitable because FSL's Program for Regional Observing and Forecasting Services (PROFS) has been conducting forecast exercises there for several years, and therefore an extensive mesoscale retrospective data set exists, as do extensive verification data. In addition, PROFS maintains a network of automated surface observing platforms (the PROFS mesonet), which provides valuable data for regional weather forecasting.

## 2. SCIENTIFIC AND TECHNICAL OBJECTIVES

SHOOTOUT-89 is an exploratory study of the efficacy of AI systems in the weather forecasting process. Data gathered during the forecast phase of the study will be used to evaluate the tasks, rules, and regulations of the study, as well as the performance of the participating systems. We expect that SHOOTOUT-89 will be followed by a series of similar studies.

We are running the AI systems on a regular basis in order to test the robustness of the systems when they are exercised daily against a wide variety of summer weather conditions. We will also compare the effectiveness of the various AI approaches, meteorological knowledge bases, and user interfaces. Some of the systems are nearly totally automated; others "collaborate" with a meteorologist, and partially serve a training and reminding function. The forecasts from the AI systems are a part of FSL's daily weather briefings; feedback from

---

the meteorologists who attend the briefings will provide additional data for future AI systems.

Although skill scores for the system will be generated, they will be useful only as a part of more extensive evaluations. For instance, lead-time, duration, size of the warned area, and need for expert human input are all important evaluation criteria that must be considered. The development of appropriate evaluation metrics is, in fact, one of the primary tasks of SHOOTOUT-89.

3. PARTICIPATING PROGRAMS

At the date of this writing (late June 1989), the following systems are participating in SHOOTOUT-89:

* Knowledge Augmented Severe Storms Predictor (KASSPr). This was developed by the Atmospheric Environment Service of Canada's Department of the Environment (Environment Canada) in cooperation with Digital Equipment Corporation. The system is a traditional expert system both in structure and in the method of creation. All knowledge is coded in rules written in OPS-5. Rule firing is controlled by the forward-chaining OPS-5 production system.

Knowledge was elicited in a series of interviews and exchanges of documentation between the developer (Bruno deLorenzis) and an expert in severe weather forecasting (John Bullas). Demonstrations of the system were combined with additional knowledge acquisition sessions. At a fairly early stage a prototype of the system was delivered to the expert for the running of historical test cases.

KASSPr requires both NGM numerical model output and extensive input from the meteorologist running the system. The meteorologist identifies and draws on the computer screen the forecasted positions of numerous meteorological features such as fronts, and pressure, thermal, and vorticity troughs and ridges (deLorenzis, 1988). After this interaction, the system generates forecasts without further intervention.

KASSPr first evaluates the meteorological situation for necessary conditions. These are relatively few but must all be met for a given area or point to warrant further consideration. Next the situation is evaluated for sufficient conditions. A certain percentage must be present for the area or point to be classified as "severe." Also for the first time a severity category and probability of occurrence is assigned. Finally, a set of modifying conditions is applied to the probabilities and the severity factors. Care must be taken in this phase to ensure independence of the order in which conditions are evaluated.

* GOPAD, developed by Kenneth Young of the University of Arizona at Tucson, was modified for use in SHOOTOUT-89 with the help of Paul Lampru of Consultant's Choice, Inc. This system uses multiple discriminant analysis to make forecasts by analogy.

In order to tune GOPAD for SHOOTOUT-89, the developers used three years of sounding (weather balloon) information and weather verification information collected in the forecast regions as input to an optimizing, statistical, analog forecasting program.

For SHOOTOUT-89, GOPAD requires PROFS mesonet data for 1200 GMT through 1555 GMT, and 1200 GMT soundings for Denver and surrounding National Weather Service (NWS) locations. For each of the forecast regions, the GOPAD model produces (1) standard forecasts, (2) the probability of tornadoes and/or funnel clouds, (3) probability distributions of expected maximum hailstone sizes, and (4) probability distributions of expected peak wind gusts.

Two versions of GOPAD are being run for this experiment. One version (called the "learning version") receives verification information for each previous forecast day in order to adjust the model on the basis of this latest information. The other "static version" receives no verification data and consequently does not change.

* CONVEX, from the NOAA/National Environmental Satellite, Data, and Information Service, is a rule-based production system that relies upon backward-chaining inferencing provided by the EXSYS expert system shell.

The development of CONVEX (Weaver and Phillips, 1987) was similar to the development of KASSPr. Weaver, who has considerable severe storm forecasting experience, was the expert who provided the rules. Phillips acted as the knowledge engineer and entered the rules into the EXSYS shell. This was an iterative process that allowed each subtopic to be separately entered into the knowledge base, and then evaluated as an integral part of the system.

CONVEX also depends a great deal on information provided to it from a sounding analysis package. The sounding analysis is accomplished through an incremental, one-dimensional updraft model of the atmosphere to which an empirically derived precipitation drag coefficient is appended. The utility of the analysis package to CONVEX is twofold. First, it determines the relative instability of the host air mass and its likelihood of initiating convection over the front-range. These are used to develop 3-9 hour regional convective weather forecasts. Second, it uses the surface mesonet temperature and dew point measurements and a linear, time-dependent, boundary layer mixing function to derive the same two parameters. This provides the basis for formulating nowcasts of severe weather probabilities within each forecast region.

Input to CONVEX consists of the Denver morning sounding and the most recently available PROFS mesonet data. Reasonably knowledgeable responses from a meteorologist to several questions regarding synoptic-scale conditions also are required. The meteorologist also is given the opportunity to override the model-derived forecasts for regional surface and dew point temperatures. No interpretation of output is necessary; probabilistic forecasts are issued for the specific forecast regions.

It is easy to have CONVEX display its rules, and thereby explain its reasoning. This has been useful for understanding the program's behavior, particularly in those situations when the forecasts of CONVEX diverge from those of most of the other programs. (It is not clear yet what effect these differences will have on relative forecast skill.)

* Additive Linear Prediction System (ALPS) was developed by Tom Stewart of the State University of New York at Albany and Cynthia Lusk of the University of Colorado at Boulder. This linear model was developed by using the methods of judgment analysis, from the field of cognitive science.

The approach is based on a well-established result in judgment and decision research: Under certain conditions, simple algebraic models can capture the skill in the judgments of an expert, and often such models can outperform the expert. In particular, weighted-sum linear models have repeatedly been shown to outperform expert judges in tasks involving high uncertainty, intercorrelated variables, and monotonic relations between variables and the observed event. This occurs because linear models provide robust prediction systems and they are perfectly consistent, whereas human judges are not. This equation was easier to develop and simpler than the other expert system models in the study in three ways: (1) the equations are based only on a subset of the variables used by the other systems; (2) the variables are combined by a simple algebraic rule rather than by complex production functions; (3) meteorological expertise in the development of the model was required only in selecting variables, estimating their relative weights, and calibrating the output.

Input consists of values of six key variables or precursors for each forecast region (positive buoyancy, wind shear, surface temperature, humidity, wind speed, and wind direction). Some of these values are determined by the operator from the Denver morning sounding; others are read automatically from the PROFS mesonet data. The output is limited to the standard probability forecasts for each region.

* WILLARD was initially developed by Steve Zubrick at the Radian Corporation (Zubrick and Riese, 1985). He is continuing the development and evaluation while working at the National Weather Service (NWS). WILLARD was developed as a prototype to forecast the potential of severe thunderstorms in the central United States; its forecasts are designed to be similar to the Convective Outlooks issued thrice daily by forecasters at the National Severe Storm Forecast Center.

The WILLARD expert system is composed of a structured hierarchy of about 30 modules, each containing a single decision rule. All decision rules within each module were developed using the inductive generalization feature (viz., ID3) of RuleMaster, an expert system shell. The rules were subsequently modified by hand. For SHOOTOUT-89, additional rules were added that pertain to shorter range severe thunderstorm forecasting as practiced by forecasters at the Denver NWS Forecast Office.

Each module contains conditions (e.g., moisture availability, wind flow) that will aid in discriminating between severe and nonsevere events. Examples of forecaster decision-making using these conditions are fed into the ID3 algorithm, and a logically consistent decision rule is generated that is examined by the developer (meteorologist) for suitability and correctness. The modular structure provides the framework for generating explanations.

WILLARD will ask the meteorologist as many as 40 questions before producing a forecast. These questions pertain to current synoptic and mesoscale features along with numerical forecast guidance. Only a few questions are asked if critical factors (such as available moisture) are not favorable for convection. Additional questions are asked if conditions for severe weather look more promising.

* Objective Convective Index (OCI) was conceived and developed by Robert Shaw with considerable input from Thomas Corona, Denice Walker, and many participants of the PROFS Real-Time 1987 forecast experiment (RT-87).

Several indices of severe weather potential were monitored during the RT-87 exercise but most focused on only a small part of the severe weather forecast problem, and none was tailored to Colorado convective climatology. The OCI draws from a wide variety of data types and sources, uses proven severe weather forecasting principles, and is specifically geared to answering the question, "What are the chances of severe weather this afternoon in the PROFS mesonet region?"

Many Boulder-area meteorologists were used as experts in the developmental effort. A list of potential predictors was compiled. The archived data were insufficient and the number of potential predictors too large for effective regression equations to be generated. Instead, predictors were subjectively weighted and a linear model was built. Heuristic rules were added to identify and account for relationships among variables that might inhibit convection. To ensure thoroughness, both observed data and NWP model forecast data are included for each of the four basic weather elements: temperature, pressure, moisture, and wind. The OCI is based on universal convective forecasting principles and therefore can be modified for use in any climate.

The meteorologist is required to input surface observations, Denver sounding data, and NGM forecast data for Denver and Cheyenne. Categorical and probabilistic forecasts for three weather categories and three regions are produced without further interaction.

The following additional systems are expected to begin to participate in SHOOTOUT-89 in the near future.

\* Severe Weather Intelligent Forecast Terminal (SWIFT), from Environment Canada, is an outgrowth of a system called METEOR, an early expert system developed for the Alberta Hail Project in 1984 (Elio and de Haan, 1985). SWIFT uses numerical weather prediction model results and weather observer reports to produce a map of potentially active convective regions.

The meteorological basis for the system is a statistical thunderstorm intensity prediction equation developed by Geoffrey Strong. The predictand is called the Synoptic Index of Convection or SC4 and is in itself a combination of four other indices, some of them statistical predictands obtained with other equations. Strong found through experience that the SC4 could best be used in combination with an index of available surface moisture, and he selected the pseudo-equivalent potential temperature. Through further experience Strong found that areas where storms would initiate and areas that were under the maximum threat could be determined algorithmically from patterns formed by contouring these two fields and noting mid-level advective wind directions. Thus the first outputs of the SWIFT system are the initiation of threat areas as determined algorithmically.

The AI portion of the system is invoked to modify the maximum intensity in each threat area. This is done by first preparing a chart depicting various cloud and precipitation patterns at the initial analysis time. The maximum threats are then modified on the basis of occurrence of certain cloud and precipitation types.

SWIFT outputs charts of the geographical area covered with forecasts of maximum daily thunderstorm severity for each region and indicates areas of likely storm initiation. SWIFT then outputs severe weather boxes with a maximum daily thunderstorm intensity as produced by the rules applied in the AI portion of the system. SWIFT also produces a verbal forecast for each severe weather box that explains why the maximum severity was modified from the algorithmic calculation.

\* Severe Weather Advisory Program (SWAP) was first developed by William Roberts (Roberts, 1988) while he was working at the National Center for Atmospheric Research. SWAP is a rule-based system using backward chaining.

SWAP has a static database made up of rules pertaining to five indices computed from the Denver morning sounding. To trigger a rule, the index value must lie in the range favoring a particular hypothesis. Three hypotheses, corresponding to the three forecast categories, were tested independently, and probabilities assigned to each hypothesis were calculated from confidence factors assigned to each triggered rule.

SWAP is a "self learning" program similar to GOPAD. SWAP uses a dynamic data base of confidence factors based on each index's past success at forecasting severe, significant, and nonsignificant weather. If an index value indicates the correct forecast on a given day, it is rewarded with additional weight. Some weight is removed if the index's forecast is not correct.

Along with probability forecasts, the program produces a histogram that is based on the confidence factors associated with each index plus an indication of which hypothesis each index validated.

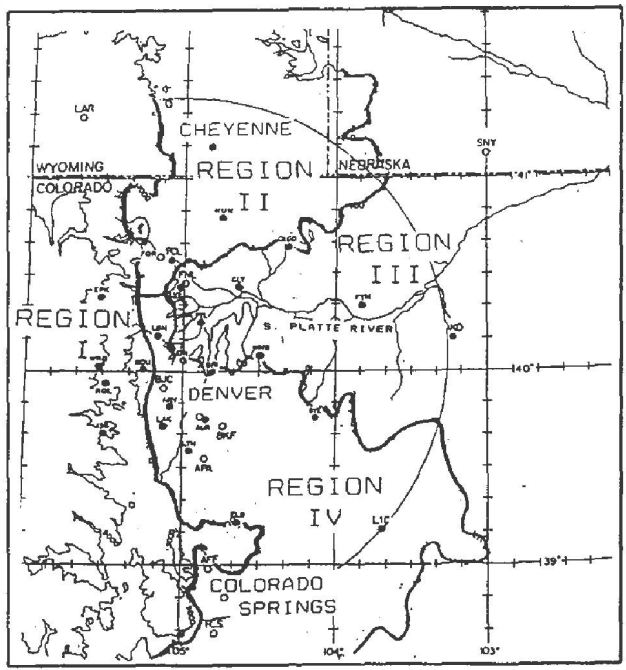

Figure 1. Forecast regions for SHOOTOUT-89.

4. EXPERIMENT DESIGN AND OPERATIONS

SHOOTOUT-89 runs from mid-May until mid-August, 1989. Each weekday of the exercise, each system (with the few exceptions noted below) generates forecasts of the mutually exclusive and exhaustive probabilities of occurrence of each of three weather categories, in each of four designated zones in northeastern Colorado (Fig. 1). The forecast zones, chosen as being climatologically distinct, are based on work by Weaver et al. (1987). The weather categories are the following:

Category 0: Nonsignificant weather, defined as the absence of category 1 or 2 weather.

Category 1: Significant weather, defined as a storm that has any of the following: a) hail with a diameter from 0.25 to less than 0.75 inches, b) surface winds between 35 kt and 49 kt, c) 2-inch-per-hour or greater rainfall rate, or d) a funnel cloud (aloft).

Category 2: Severe weather, defined as a storm observed to have any one of the following: a) hail with diameter equal to or greater than 0.75 inches, b) surface winds of 50 kt or greater, or, c) a tornado.

There are two exceptions to these forecast requirements: (1) WILLARD generates forecasts only for the overall forecast region, not for the individual regions shown in Fig. 1. (2) OCI does not generate forecasts for region 1.

Data to be used by the programs first become available at about 9 a.m. local time; however, the bulk of the needed data is not available until about 10:15. Forecasts from all systems except OCI are finished by approximately 11:15 (1715 GMT). Because of the amount of manual input required, OCI is run in the early afternoon (but using morning data). The valid time for the forecasts is 1900 GMT to 0200 GMT (1 p.m. to 8 p.m. MDT).

At approximately 11:25 a.m., forecast results from the AI systems are presented orally and with overhead transparencies at the FSL daily weather briefing. This briefing is led by one of the many research meteorologists in Boulder, and is generally attended by 20 to 30 research meteorologists. At this time verification data from the previous forecast day are also presented.

5. EVALUATION

5.1 Forecast skill evaluation

As discussed in section 2, a primary objective of SHOOTOUT-89 is to compare the strengths of various AI approaches. Meteorological truth data are needed in order to verify and evaluate the systems' forecasts.

A full-time verification coordinator (VC) is responsible for gathering and documenting verification data for SHOOTOUT-89. (The data are also used by the Denver Weather Service Forecast Office (WSFO) in its own forecast verification studies.) During times of expected significant or severe weather, the VC is stationed at the Denver WSFO, where there is access to real-time radar data. When radar (or other) data suggest possible significant or severe weather, the VC commences calling cooperating observers in potentially affected regions. The VC is also available to receive reports phoned in to the WSFO, and on days following possible weather events, makes follow-up phone calls.

The following sources of verification data are now, or soon will be, available:

o   A volunteer spotter network and a paid cooperative observer network are sponsored by NWS.

o   Police and fire stations, county emergency preparedness staffs, and highway road crews are being contacted, and arrangements are being made for them to respond to telephone queries regarding possible significant and severe weather events.

o   A network of amateur radio operators provides verbal (and shortly, video) weather information during times of expected severe weather.

o   Weather service offices in Colorado Springs and Cheyenne provide information and respond to telephone queries.

o   Automated mesonet stations provide data on maximum wind gusts and rainfall rate. Such data have good time continuity, but are available at only relatively few points in the forecast area.

o   Daily weather observations are recorded by a network of approximately 30 weather observers recruited especially for SHOOTOUT-89. Observations are mailed in monthly.

o   Occasionally, volunteer chase teams operate. These generally consist of research meteorologists who maintain contact with the VC by cellular phones, or who report the results of their observations after the fact.

It should be noted that the verification data are unlikely to be robust enough for us to make valid distinctions between significant and severe weather. We asked for these two weather categories primarily to allow for system intercomparison. Similarly, because of the difficulty of gathering verification data in the mountains, it is unlikely that we will be able to generate valid skill scores for Zone I. Again, we have included this zone primarily to facilitate system intercomparison.

We plan to analyze the forecast-observation data from a number of viewpoints in order to assess more than one characteristic of the AI systems. We have agreed that our analyses will not attempt to declare any system an overall winner.

We have calculated Brier Skill Scores for the first 32 days of the experiment. (For these skill scores, we have compared the forecasts of the systems with forecasts made by simply predicting a probability of significant or severe weather equal to the summer climatological base rate.) Two of the six systems are currently showing skill up to 36% better than that of climatology. The least skillful system has skill 50% less than that of climatology. There is considerable regional variation in forecast skill; all systems are least skillful over the region that includes metropolitan Denver.

Because of the difficulties of gathering reliable verification data, because of the exploratory nature of SHOOTOUT-89, and its emphasis on interpretive intercomparisons, we are not performing comparisons between the skill scores from the AI systems and those from operational forecasts.

### 5.2 Non-skill evaluation

A full analysis of the performance of the systems requires that we evaluate the systems on metrics beyond forecast skill. Our ultimate goal is to provide useful feedback to the system designers about how their systems facilitate the overall forecast process. To that end, we are addressing the following questions.

o   How "task intensive" is each system? That is, how much time is required to run each, what data are necessary for input, and how much meteorological expertise is required to operate each system?

o   How much system-specific training is required to operate each system?

o   Does an "operator effect" exist for the systems? That is, do operators using the same meteorological data generate different forecasts?

o   How much effort was necessary in order to "tune" each system to Colorado weather; and how much effort would likely be required to tune the systems for other locations?

o   How satisfied are meteorologists with each system? In particular, are the required inputs relevant and non-redundant from a meteorological perspective, and are the outputs useful in generating forecasts?

### 6. SUMMARY

The goal of SHOOTOUT-89 is to provide AI system designers with feedback with which they can design systems that will contribute usefully to the overall forecast process. We will present some initial evaluation results at the Workshop.

### 7. ACKNOWLEDGEMENTS

We would like to thank the following people, without whose help, SHOOTOUT-89 would not exist. For having the idea: J. Carr McLeod of Environment Canada; for gathering verification data: E. Ellison; for providing system support: D. Davis, M. Govett and R. Lipschutz.

PROFS graciously provided the meteorological data used by the systems; the Denver Forecast Office of the National Weather Service provided considerable support in the gathering of verification data.

SHOOTOUT-89 was partially funded by a grant from the director of NOAA's Environmental Research Laboratories, Dr. J.O. Fletcher.